\documentclass[nohyperref]{article}

\usepackage{microtype}
\usepackage{graphicx}
\usepackage{subfigure}
\usepackage{booktabs} % for professional tables

\usepackage{hyperref}

\usepackage[accepted]{icml2022}

\usepackage{amsmath}
\usepackage{amssymb}
\usepackage{mathtools}
\usepackage{amsthm}

\usepackage[capitalize,noabbrev]{cleveref}

\theoremstyle{plain}

\theoremstyle{definition}

\theoremstyle{remark}

\usepackage[textsize=tiny]{todonotes}

\usepackage{comment}

\icmltitlerunning{MaiT: Leverage Attention Masks for More Efficient Image Transformers}

\begin{document}

\twocolumn[
\icmltitle{MaiT: Leverage Attention Masks for More Efficient Image Transformers  
           }



\icmlsetsymbol{equal}{*}

\begin{icmlauthorlist}
\icmlauthor{Ling Li}{comp}
\icmlauthor{Ali Shafiee Ardestani}{comp}
\icmlauthor{Joseph Hassoun}{comp}
\end{icmlauthorlist}

\icmlaffiliation{comp}{Samsung Semiconductor Inc., San Jose, CA, USA}

\icmlcorrespondingauthor{Ling Li}{ling.li@samsung.com}

\icmlkeywords{Computer Vision, Deep Learning, image classification, Vision transformers}

\vskip 0.3in
]



\printAffiliationsAndNotice{}  % leave blank if no need to mention equal contribution


\begin{abstract}
Though image transformers have shown competitive results with convolutional neural networks in computer vision tasks, lacking inductive biases such as locality still poses problems in terms of model efficiency especially for embedded applications. In this work, we address this issue by introducing attention masks to incorporate spatial locality into self-attention heads. Local dependencies are captured efficiently with masked attention heads along with global dependencies captured by unmasked attention heads. With Masked attention image Transformer – MaiT, top-1 accuracy increases by up to 1.7\% compared to CaiT with fewer parameters/FLOPs, and the throughput improves by up to 1.5$\times$ compared to Swin. Encoding locality with attention masks is model agnostic, and thus it applies to monolithic, hierarchical, or other novel transformer architectures. 
\end{abstract}

\section{Introduction}
\label{submission}

Convolutional neural networks (CNNs) \citep{alexnet, resnet, efficientnet} have been the de facto model for computer vision (CV) tasks, which are inherently equipped with inductive biases such as translation equivariance and locality. Recently, vision transformers \citep{ViT, deit} are gaining momentum translating the success from transformer-based models in natural language processing (NLP) tasks \citep{transformer, bert}. Self-attention heads excel at capturing the long-range dependencies in sequences but struggle at focusing on local information. Unlike CNNs which have gone through many iterations of optimization, vision transformers are not very efficient especially for embedded applications \cite{mobilevit}.

Naturally, combining the benefits of both CNNs and vision transformers is promising to further boost the performances of CV models. The question remains how to effectively integrate inductive bias such as spatial locality into transformers. One direction is to utilize convolutional blocks to extract spatial information by adapting either the patch-token embedding layer, self-attention module, or feed-forward layers, to form CNN-transformer hybrid structures \cite{localvit, botnet,levit,cvt, ceit, cmt}. 
However, forcefully inserting convolutional operations into transformers may potentially constrain the learning capacity of transformers.

To capture the spatial information without significantly changing the transformer model architecture, Chu et al. \yrcite{cvpt} introduce extra positional encoding. Han et al. \yrcite{tnt} and Chen et al. \yrcite{crossvit} fuse local and global representations using multiple transformer blocks or branches to simultaneously process images at different scales such as pixel-level, small-patch, or large patch. Yuan et al. \yrcite{t2t} apply a layerwise tokens-to-tokens transformation to capture local structure. These approaches usually come with the cost of extra parameters and model complexity, thus potentially lowering the inference speed.

d'Ascoli et al. \yrcite{convit} and Zhou et al. \yrcite{deepvit} improve the self-attention for better representation with gated positional self-attention and learnable transformation matrix respectively. Hu et al. \yrcite{lrnet} and Ramachandran et al. \yrcite{stand-alone-attention} adapt the self-attention module and improve the performance of CNN models. Liu et al. propose Swin \yrcite{swin} to capture locality within shifted windows in a hierarchical structure. Though it could save computation in some cases, the added model complexity lowers its throughput compared to similar sized DeiT \cite{swin}.   

In contrast to prior works, we propose the use of attention masks to efficiently incorporate spatial locality, without adding extra parameters, FLOPs, or model complexity. The attention mask approach is model agnostic and thus can be applied to either monolithic, hierarchical, or other novel transformer architectures.
%We apply attention masks on attention heads to effectively extract local dependencies, by allowing information aggregation only from the closest neighbors. 
By applying attention masks, local dependencies are extracted effectively through information aggregation from the closest neighbors. 
This liberates other unmasked heads to learn global information more smoothly. 
We name the model, Masked attention image Transformer (MaiT), which is built on top of DeiT \citep{deit}. MaiT gathers both local and global information at the same time from different heads.
 
%the regularization effects from attention masks facilitate the training of deep transformers by guiding the attention map learning and promoting diversity across transformer layers.

We proved that less is more in this specific case with attention masks. MaiT achieves up to 2.5\% higher top-1 accuracy on ImageNet \citep{imagenet} with the same model architecture as DeiT \citep{deit}. Additionally, deep MaiT outperforms CaiT \citep{cait} by up to 1.7\% in top-1 accuracy with fewer parameters/FLOPs. Moreover, MaiT improves the throughput by up to 1.5$\times$ compared to Swin due to the structural simplicity. 

In summary, we make three major contributions in this work: 1). We propose attention masks to improve model efficiency by encoding the spatial locality into self-attention heads, 
%without adding extra parameters/FLOPs. 
which reduces the computation complexity of attention maps from quadratic to linear. 2). We present a quick empirical search strategy for the masking scheme exploration, along with an automatic end-to-end search alternative. 3). We demonstrate the validity of MaiT in both monolithic and hierarchical architectures with competitive accuracy and throughput. 
%We also reveal the importance of the locality across layers and the performance impact of attention masks on deep vision transformers. 
The attention mask is applicable to other vision transformers, not limited to the models presented in this work. In addition, attention masks can be easily combined with many other optimization techniques to further boost vision transformer performances.   

\section{Related work}

%Spatial locality is an integral part of the convolutional operation with weight filters attending to local regions of input feature maps. 
Vision transformer (ViT) \citep{ViT} is the first pure transformer-based model on vision tasks, but it requires a large private labeled dataset JFT300M \citep{jft} to achieve competitive performances. Data-efficient image transformer (DeiT) \citep{deit} improves upon ViT models by introducing stronger data augmentation, regularization, and knowledge distillation. Class-attention in image transformer (CaiT) \citep{cait} extends DeiT by increasing the number of transformer layers. To overcome the difficulties of training deeper transformers, CaiT introduces LayerScale and class-attention layers, which increase the parameters and model complexity. 

Various approaches have been proposed to capture local and global features more efficiently in vision transformers. Tokens-to-Token vision transformer (T2T) \citep{t2t} proposes an image transformation by the recursive token aggregation to capture local structure.  
 Stand-alone self-attention \citep{stand-alone-attention} applies local self-attention layer to replace spatial convolution and outperform original ResNet models. Even though sharing value and key spatially is parameter efficient in this approach, content-based information is lost.
 Transformer-iN-Transformer (TNT) \citep{tnt} models both patch-level and pixel-level representations and applies outer and inner transformer blocks to extract global and local information respectively.
ConViT \citep{convit} proposes the gated positional self-attention to incorporate soft convolutional biases.
CrossViT \citep{crossvit} proposes a dual-branch transformer architecture for multi-scale feature extraction. 

The hierarchical structure is well adapted for pixel-level prediction tasks such as semantic segmentation and object detection. Pyramid Vision Transformer (PVT) \citep{pvt} introduces a progressive shrinking pyramid and spatial-reduction attention with fine-grained image patches. Swin Transformer \citep{swin} applies a hierarchical structure with shifted windows of varying sizes. Twins \citep{twins} deploys interleaved locally-grouped self-attention and global sub-sample attention layers to improve performances. 

To optimize transformer and save computation, Wu et al. \yrcite{centroid} uses centroid attention to extract and compress input information,  Jaegle et al. \yrcite{perceiver} iteratively distill inputs into latent space with attention bottlenecks, Wang et al. \yrcite{dynamic} dynamically adjust the number of tokens with multiple cascading transformers, and Wu et al. \yrcite{visual} introduced semantic token to replace pixel-based transformers to save computation. Masking has been used in NLP tasks as a sparsification method to reduce the computation complexity, as well as to capture local information \cite{startransformer, sparse, longformer, etc}. 

There are hybrid architectures fusing convolutional and transformer blocks, such as LocalViT \citep{localvit}, BoTNet \citep{botnet}, LeViT \citep{levit}, BossNet \citep{bossnet}, CvT \citep{cvt}, CoaT \citep{coat}, CMT \citep{cmt} for higher accuracy and faster inference. %LocalViT \citep{localvit} introducing locality by adding depth-wise convolution into the feed-forward network. 

Unlike prior literature, our work explores the intrinsic capability of pure transformer block on incorporating spatial locality for both monolithic and hierarchical structures. 
%and the impact of the locality along the depth direction. 
This work is also inspired by the emerging graph attention network \citep{gat}, borrowing the concept of message passing and information aggregation from nearest neighbors. 

\section{Vision Transformer Preliminaries}
\label{headings}
The transformer architecture introduced by Vaswani \citep{transformer} inspired many model variants with remarkable success in NLP tasks. ViT \citep{ViT} extends pure transformer-based architecture into CV applications. Instead of pixel-level processing, ViT splits the original images into a sequence of patches as inputs and transforms them into patch tokens, for better computation efficiency. In general, ViT consists of 3 fundamental modules: embedding layer, multi-head self-attention, and feed-forward network.

To process images in transformer, the original  RGB images (224$\times$224) are flattened into a sequence of $\it N$ (14$\times$14) patches. Each patch has a fixed size (typical 16x16 pixels). Patches are then transformed into patch embedding with hidden dimensions ($\it D$) of 192, 384, and 768 for tiny, small, and base models respectively in ViT/DeiT. In addition to patch tokens, the embedding layer also integrates positional information, classification, and knowledge distillation through the positional token, class token, and distillation token, respectively. 

%Positional token is added into the patch embedding with a trainable positional embedding. However, this positional embedding is added only in the embedding layers. The spatial information is largely lost in the transformer layers since all-to-all attention is invariant to the order of the patches.

%The class token is another trainable vector (1$\times \it D$), concatenated to the patch tokens (total $\it N$+1). It is used to collect information from the patch tokens to make output predictions, while also spreading information among patch tokens during training. 

%Distillation token is sometimes added for knowledge transfer from teacher models, such as a CNN model. When training the distilled version of the model, a distillation token is further concatenated to the patch token along with the class token (total $\it N$+2). %The distillation token is complementary to the class token by providing extra information from the teacher model. At test time, both class or the distillation token, or fusion of the two, are inputs to the linear classifier. 

Multi-head self-attention (MHA) module has multiple parallel attention heads, where each of them comprises three main components: Key ($\it K$), Query ($\it Q$), and Value ($\it V$), Key and Query are trained and multiplied to estimate how much weights on each corresponding token in Value for attention (Attn) output:
\begin{equation} \label{attention}
    \text{Attn}(K, Q, V) = Softmax\left( \frac {QK^T} {\sqrt{d}} \right) V
\end{equation}
Where softmax is applied to each row of the input product matrix ($QK^T$) and $ \sqrt{d} $ provides appropriate normalization.     
Multiple attention heads in MHA attend to different parts of the input simultaneously. Considering $\it H$ heads in MHA layer, the hidden dimension $\it D$ is split equally across all heads ($\it D=H\times d$). 

Feed-forward network (FFN) follows after the MHA module, containing two linear transformation layers separated by GeLU activation. 
Both MHA and FFN use skip-connections with layer normalization as the residual operation. 

\section{Masked Attention Head}
\label{others}

Spatial locality plays a crucial role in computer vision tasks. CNN models capture it efficiently using the sliding filter of shared weights, typically with a receptive field of 3$\times$3, 5$\times$5, or 7$\times$7. In contrast to CNN models, the locality is not explicitly introduced in the transformer structure.  
With masked attention heads, we can specifically insert locality into self-attention modules without much overhead. The key idea is to apply a mask on the all-to-all attention products (i.e. $QK^T$) to reinforce the weights from the closest neighbors by aggregating information only from tokens selected by the mask. Meanwhile, the unmasked heads retain the capability to extract global features effectively. 

\begin{figure}[ht]
\vskip 0.1in
\begin{center}
\centerline{\includegraphics[width=\columnwidth]{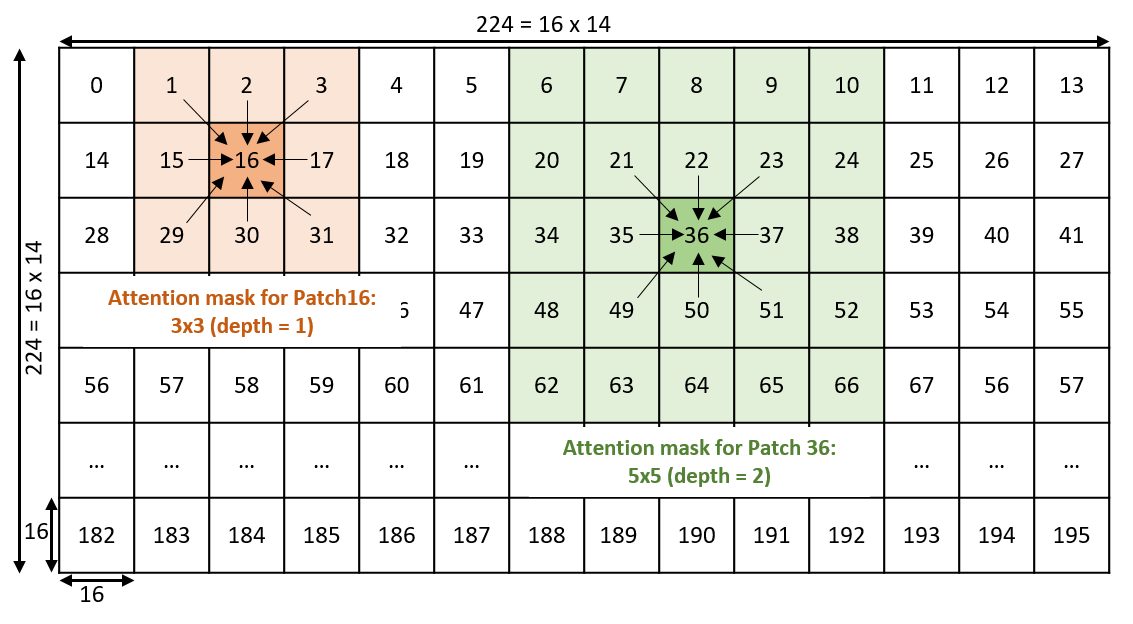}}
\caption{The illustration of 3$\times$3 (orange) and 5$\times$5 (green) attention masks.}
\label{attn mask}
\end{center}
\vskip -0.1in
\end{figure}

Figure~\ref{attn mask} illustrates an example of our proposed $R \times R$ attention masks, where $R$ is the mask width/height. The orange box shows a 3$\times$3 mask, where only the direct neighboring patches are selected. Specifically, Patch 16 only gathers information from the closest neighbors of Patch 1, 2, 3, 15, 17, 29, 30, and 31, and ignores the rest of the patches. 
This is different from the typical all-to-all attention, where  Patch 16 attends to all 0-195 patches. 
We can easily expand the depth of the attention mask beyond the closest neighbors, to second-level neighbors (green box in Figure~\ref{attn mask}) or more. Note that the class token still attends to all the patches to collect and spread information during forward and backward passes.
Since each attention product selected by the mask is calculated by $\it Q$ and $\it K$, the masked attention head also retains the content-based information. 

The attention mask is added before Softmax, regulating the distribution of attention maps to focus more on the closest neighbors. The masked attention (M-Attn) is computed as: 
\begin{equation} \label{masked attention}
     \text{M-Attn}(K, Q, V) = Softmax\left( \frac {M \odot QK^T} {\sqrt{d}} \right) V
\end{equation}

where $M \in \{0, 1\}^{(N+1) \times (N+1)}$ is a binary attention mask, encoding the spatial locality into the attention head by passing through only the weights from close neighbors and setting the rest to zero. $N+1$ is the number of patch tokens plus the class token. 

Note it's important to add the mask before the softmax operation because it allows the model to learn the importance of the locality flexibly. More precisely, unselected patches appear as $e^0=1$ in the numerator of the softmax operation. 
Thus, if the attention product result of the closest neighbors is meaningfully larger than zero (i.e. $M \odot QK^T \gg 0$), it suggests that local information dominates. 
However, if those results are negative or close to zero, it implies that local information is insignificant and global information is more important.
\emph{Therefore, inserting masks before the softmax operation allows attention heads to enforce locality or bypass it.} Using a combination of masked and unmasked heads in each layer enables the transformer to exploit both local and global information more efficiently. 

\subsection{Computation Complexity}
One emerging issue with transformer models is the poor scalability to compute the attention map, which increases quadratically with respect to the number of input patch tokens. This severely hinders its applications such as high-resolution image processing. Attention masks effectively reduce the computation complexity from quadratic $O(N^2)$ to linear $O(N)$. This approach scales even better than the windows-based self-attention proposed in Swin \cite{swin}. More specifically, the complexities of various methods are listed in Table~\ref{computation complexity}. Note that Swin adds model complexity to capture global information across layers which lowers its throughput compared to MaiT as shown in Section 5. 

The computation saving is substantial in cases with high-resolution input images or hierarchical structures where earlier layers contain a significantly large number of patch tokens due to small patch sizes. For example, if the patch size is 4x4 (same as the first stage of Swin), which translates to a total  (224/4)$^2$ or 3136 patch tokens. As a result, calculating the attention map consume 84.5\% of the total computation in a transformer block (2.23 GFLOPs). On the other hand, if applying masked attention heads to this block, the computation for the attention map reduces to 9/3136 or 0.3\% of the original value. This reduces the computation of the entire transformer block to 0.35 GFLOPs, an 84.2\% reduction. 

\begin{table}[t]
\caption{The computation complexity of attention maps from various methods: W-MHA is the windows-based attention \cite{swin} where $M$ is the window size, typically as 7; M-MHA is the masked attention, where R is the mask width/height, set as 3 in this work.}
\label{computation complexity}
\vskip 0.1in
\begin{center}
\begin{small}
\begin{sc}
\begin{tabular}{lccc}
%\toprule
\multicolumn{1}{c}{\bf Methods}  &\multicolumn{1}{c}{\bf MHA} &\multicolumn{1}{c}{\bf W-MHA} &\multicolumn{1}{c}{\bf M-MHA} \\
%\hline 
\midrule
Complexity          &$2N^2D$        &$2(M^2N)D$      &$2(R^2N)D$ \\
%\bottomrule
\end{tabular}
\end{sc}
\end{small}
\end{center}
\vskip -0.1in
\end{table}

\begin{figure}[ht]
\vskip 0.1in
\begin{center}
\centerline{\includegraphics[scale=0.5]{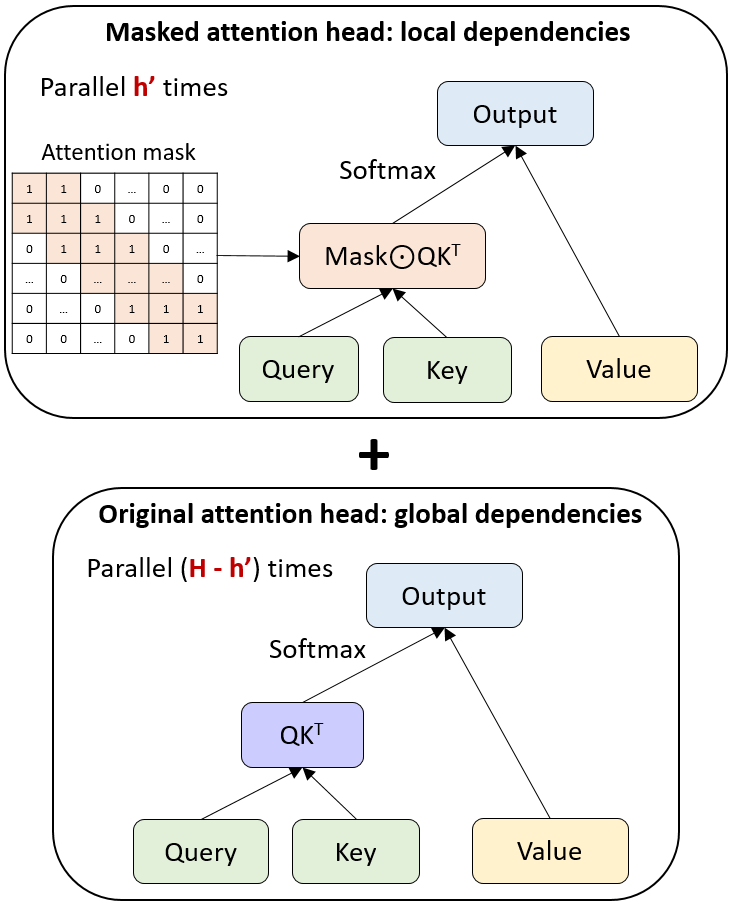}}
\caption{Masked attention heads and original attention heads to capture local and global dependencies respectively.}
\label{attn heads}
\end{center}
\vskip -0.1in
\end{figure}

\subsection{Attention Locality Score (ALS)}
The softmax operation transfers the results of $QK^T$ into the probability space. Thereby each row of the attention map ($A$) sums up to 1. For each patch token, the probability of focusing on local neighbors equals the sum of their neighbors' attention map weights. 
If this number approaches one, local information is crucial; whereas if it approaches zero, global information matters more. 
For Patch Token $n$, we define $ALS_n$ as its Attention Locality Score (ALS). The mean of all $\it N$ patches is the ALS for each attention head, as follow:
\begin{equation} \label{ALS definition}
    ALS = \frac {\sum_n ALS_n} {N}, \text{where } ALS_n = \sum_i^{N} (M \odot A)_{n, i} 
\end{equation}
where $M$ is the same attention mask defined earlier, and $A = Softmax\left( M \odot QK^T / \sqrt{d} \right) $ is the attention map, or $ A = Softmax\left( QK^T /  \sqrt{d} \right) $ for unmasked heads, and $n, i$ denotes the row and column index of the attention map respectively. 
We later use the ALS metric to get insights about the impact of different attention heads at various layers and guide the attention mask placement.

\subsection{Masking Strategy}
With total $H$ attention heads, $h^\prime$ number of attention heads can be allocated to focus on local information. The rest of the unaltered attention heads $(H - h^\prime)$ captures global dependencies. Therefore, we can extract the local information through masked attention heads and global information through original attention heads at the same time, as shown in Figure \ref{attn heads}. 

Besides the size of the attention mask, the number of masked attention heads in the MHA module and the position (which layer to insert the mask) are also hyperparameters. 
Though masks encode locality into the attention heads, the regularization from masking (pruning attention map) can also limit the learning capacity. 
Consequently, where to add attention masks requires careful consideration. However, the search complexity is easily up to $2^{36}$ for a 12-layer model with 3 heads, and worse for larger models.

Thus we provide two distinct approaches for a quick empirical search and end-to-end automatic search. 

\textbf{Quick empirical search} To obtain a decent mask placement within 2-3 training iterations, we leverage the insights from ALS study in Section 5.3 and follow a masking strategy below:

\begin{enumerate}
    \item \textbf{Initialization}: add the attention mask to only one head (Head 0) for all layers to explicitly introduce locality in every layer and form a local path from the beginning to the end
    \item \textbf{Assignment}: compute $ALS^{0,l}$ for every layer $l$ 
    \begin{enumerate}
        \item If $ALS^{0,l}$ is above 0.65: add masks to all attention heads of layer $l$, proceed to 3
        \item If $ALS^{0,l}$ is between 0.35 to 0.65: keep the mask, add no more 
        \item If $ALS^{0,l}$ is below 0.35: remove the mask 
    \end{enumerate}
    \item \textbf{Calibration}: following 2.a), compute $ALS^{h,l}$ for each head and remove the mask if $ALS^{h,l}$ is below 0.35
\end{enumerate}

Note that optimal masking is a large space as opposed to a single unique placement. The transformer model is capable of adjusting to different masking schemes flexibly. The thresholds are hyperparameters and can be used to adjust the total number of masked heads. Step 2 (a) relies on the observations that early layers (typically with ALS above 0.65) focus heavily on local features and thus more masked heads are beneficial. Step 3 will remove the redundant mask accordingly. 

\textbf{Soft Masking}
To alleviate the complicated searching problem and automatically learn the mask placement, we introduce a learnable scale factor $\alpha \in (0, 1)$ for each attention head. These scale factors are trained end-to-end simultaneously with the transformer model. We name this approach Soft Masking, which replaces the zeros in the original binary masks with scale factors. The binary masking approach is named Hard Masking. 

This scale factor penalizes the attention weights from non-neighboring patches. It allows non-local patch tokens to contribute differently. When $\alpha$ approaches 0, this attention head focuses more on local information. Otherwise when it's close to 1, the attention head attends global information. As a result, each attention head at every layer is able to flexibly determine the importance of locality. 
Note Soft Masking enriches the model capacity whereas Hard Masking enables more efficient inference. 
%Note, this introduces negligible extra parameters during training. For example, it only introduces 36 extra parameters for a 12-layer MaiT tiny model with 3 heads.

\begin{table}[b]
\caption{Details of MaiT model parameters, where D is the hidden dimension and H is the number of attention heads.}
\label{MaiT-dim}
\vskip 0.1in
\begin{center}
\begin{small}
\begin{sc}
%\resizebox{\columnwidth}{!}{
\begin{tabular}{lcccc}
%\toprule
%Model & Extra Tiny(XT) &Tiny(T) &Extra Small(XS) & Small(S)
\multicolumn{1}{l}{\bf}  &\multicolumn{1}{c}{\bf X-Tiny(XT)} &\multicolumn{1}{c}{\bf Tiny(T)} &\multicolumn{1}{c}{\bf X-Small(XS)} &\multicolumn{1}{c}{\bf Small(S)} \\
%\hline
\midrule
D    &144      &192    &288    &384\\
H    &3        &3      &3      &6\\
%\bottomrule
\end{tabular}
%}
\end{sc}
\end{small}
\end{center}
\vskip -0.1in
\end{table}

\section{Experiments}

In this section, we report the results for MaiT, implemented by applying attention masks on both the monolithic structure same as DeiT and a hierarchical structure similar to Swin. We analyze the impact of different hyper parameters including the mask size, masking strategy, and the number of transformer layers. We also evaluate the performance improvement with attention masks on various model sizes and architectures. 
We merely focus on models with fewer than 50M parameters, which are more applicable to embedded systems and mobile devices.

\subsection{Setup}
We follow the same training procedure as DeiT and Swin for the monolithic and hierarchical structures respectively, unless specified otherwise. The implementation is based on the Timm library \citep{timm}. All models are trained with ILSVRC-2012 ImageNet dataset \citep{imagenet} with input image size of 224$\times$224. Monolithic models use a default batch size of 1024 on 4 DGX A100 GPUs for 300 epochs for 12-layer models and 400 epochs for models with more than 12 layers. Hierarchical models are trained with a batch size of 512 on 8 HGX A100 GPU for 300 epochs. Model parameters we adopted are summarized in Table~\ref{MaiT-dim}.

\subsection{Impact of the mask size}

Intuitively, the field of view is larger with larger mask sizes, to gather broader neighborhood information. However, in the extreme case with mask depth of $N$+1 (standard self-attention), all the patch token information is merged and positional information is lost. Meanwhile considering attention mask as a structured pruning, smaller mask sizes translate to fewer FLOPs and more computation savings. %which might be useful for some embedded hardware devices.

To study the impact of the mask size, we mask only one attention head every layer on tiny and small DeiT models, with mask sizes of 3$\times$3 %3$\times$5, 
and 5$\times$5. As shown in Table~\ref{MaiT-depth}, different mask sizes only lead to marginal differences -0.1 - 0.4 \% in accuracy, compared to DeiT baseline.
Since each patch (16x16 pixels) is a processed sub-image and already contains some locality information, further away patch tokens barely provide extra information. Therefore, the default mask size is 3x3 in the subsequent sections. 

\begin{table}[b]
\caption{Top-1 accuracy on ImageNet for DeiT with no masking and with one masked head of mask sizes $3\times3$ and $5\times5$ for all 12 layers. MaiT adopts Soft Masking with $3\times3$ mask size and is trained for 400 epochs.}
\label{MaiT-depth}
\vskip 0.1in
\begin{center}
\begin{small}
\begin{sc}
\begin{tabular}{lcccc}
 &\multicolumn{1}{c}{\bf DeiT} &\multicolumn{1}{c}{\bf 3$\times$3} &\multicolumn{1}{c}{\bf 5$\times$5}  &\multicolumn{1}{c}{\bf MaiT}\\ 
%\hline 
\midrule
Tiny     &72.2   &72.6(+0.4)  &72.1(-0.1)  &74.7(+2.5)\\
Small   &79.8   &79.9(+0.1)  &80.0(+0.2)  &80.9(+1.1)\\
\end{tabular}
\end{sc}
\end{small}
\end{center}
\vskip -0.1in
\end{table}

In contrast to DeiT, whose accuracy saturates beyond 300 epochs \cite{deit}, MaiT-S obtains another 1.1\% gain when trained for 400 epochs. 
Accordingly, we suspect a higher learning rate and lower drop path rate are beneficial for MaiT-T since attention masks already provide some regulation. 
Adopting 0 drop path rate and 2$\times$ learning rate, we increase the accuracy by 2.5\% over DeiT-T. 

\subsection{Various Masking Strategies}
We compare the performances of 24-layer MaiT models following three representative masking schemes (Sch.), to quantify the impact of different masking strategies. Sch.1 is a naive way to combine local and global information at every layer. Sch.2 results from the quick empirical search of Hard Masking from Section 4.2 and Sch.3 is based on Soft Masking. The details of the mask placement for each scheme are shown below:    

\textbf{Sch.1}: One masked head for each layer across all 24 layers 

\textbf{Sch.2}: $H$-1 masked heads for Layer 0-7, one masked head for Layer 9-19, and no mask for Layer 20-23

\textbf{Sch.3}: Soft Masking for Layer 0-20, no mask for Layer 21-23

\begin{table}[t]
\caption{Top-1 accuracy on ImageNet for 24-layer DeiT and MaiT with different masking schemes. Sch.1 adopts only one masked head. Sch.2 applies mixed masking (Hard Masking) and Sch.3 uses Soft Masking.}
\label{masking-schemes}
\vskip 0.1in
\begin{center}
\begin{small}
\begin{sc}
%\resizebox{\columnwidth}{!}{
\begin{tabular}{lcccc}
%\multicolumn{1}{l}{\bf}  &\multicolumn{1}{l}{\bf}  \multicolumn{3}{c}{\bf MaiT} \\
\multicolumn{1}{c}{\bf } &\multicolumn{1}{c}{\bf DeiT} &\multicolumn{1}{c}{\bf MaiT(Sch.1)} &\multicolumn{1}{c}{\bf MaiT(Sch.2)} &\multicolumn{1}{c}{\bf MaiT(Sch.3)} \\
\midrule
T     &78.3     &79.1(+0.8)    &79.1(+0.8)   &79.3(+1.0) \\
XS    &80.8     &81.0(+0.2)    &81.4(+0.6)   &81.3(+0.5) \\
%\bottomrule
\end{tabular}
%}
\end{sc}
\end{small}
\end{center}
\vskip -0.1in
\end{table}

As shown in Table~\ref{masking-schemes}, top-1 accuracy improves by 0.8\% simply through adopting one masked head for MaiT-T. Soft Masking further increases the accuracy by 0.2\% for MaiT-T. MaiT-XS obtains an extra 0.4\% improvement with mixed masking Sch.2 on top of one masked head Sch.1.  Note that the initialization of the scale factor in Soft Masking for MaiT-XS is subject to further tuning. Empirically we find masking Sch.2 yields decent performances by masking heavily in the first one-third of layers to extract local features, only one mask for the middle layers, and no masking for the last few layers to shift focus more on global information. 

\subsection{Monolithic vs. Hierarchical Structures}

Monolithic structure such as ViT/Deit preserves a simpler structure and one-to-one token correspondence throughout all layers. This is beneficial for some unsupervised learning frameworks \cite{ssl} to learn a good semantic correspondence, whereas hierarchical transformers such as Swin lose this one-to-one token correspondence due to token merging along the depth direction. Nonetheless, the hierarchical structure is better suited for pixel-level dense prediction tasks such as semantic segmentation and object detection. Therefore, a model-agnostic approach to incorporate spatial locality into transformers presents wider applicability.  

\textbf{Monolithic MaiT} Comparing monolithic structured MaiT with hierarchical structured Swin, the accuracy of MaiT-XS is 0.5\% higher than Swin-T, and MaiT-S is 0.2\% less than Swin-S (Table \ref{h-mait}). Note these are achieved with 4.5M and 6.7M fewer parameters. More importantly, the simplicity of this approach with attention masks and the monolithic structure translates to 1.2$\times$ and 1.5$\times$ higher throughput with MaiT-XS and MaiT-S than Swin-T and Swin-S respectively. Besides, the Hard Masking used in MaiT presents acceleration opportunities to further boost the throughput, if given the structured sparsity support from hardware.   

\textbf{Hierarchical MaiT} To further validate the efficiency of attention masks on hierarchical structures, we constructed tiny and small hierarchical MaiT (H-MaiT) with the same architecture configurations as Swin, such as hidden dimensions, stages, depths, and the number of heads in each layer. This ensures the same total number of parameters between H-MaiT and Swin. Different from Swin, H-MaiT uses masked attention instead of window-based attention. To obtain a similar total number of FLOPs with Swin for a fair comparison, the number of masked attention heads in the 4 stages is [3, 6, 4, 0], out of [3, 6, 12, 24] total heads in each stage. 
Using the same training schedule as Swin without extensive hyper-parameter tuning, H-MaiT outperforms Swin by 0.4\% and 0.2\% for tiny and small models respectively. 

\begin{table}[t]
\caption{Performance comparison between Swin \cite{swin}, MaiT, and H-MaiT. MaiT in this table applies masking Sch.2 with LayerScale \cite{cait}. H-MaiT adopts the same architecture parameters as Swin. Throughput is measured using one 40GB DGX A100 GPU with a batch size of 64, following the same procedure in Swin \cite{deit, swin}. H-MaiT requires hardware support to exploit the sparsity in attention map computation and thus we did not include the throughput here.}
\label{h-mait}
\vskip 0.1in
\begin{center}
\begin{small}
\begin{sc}
%\resizebox{\columnwidth}{!}{
\begin{tabular}{llllc}
%\multicolumn{1}{l}{\bf}  &\multicolumn{1}{l}{\bf}  \multicolumn{3}{c}{\bf MaiT} \\
\multicolumn{1}{c}{\bf } &\multicolumn{1}{p{1cm}}{\bf Param. (M)} &\multicolumn{1}{p{1cm}}{\bf FLOPs (G)} &\multicolumn{1}{p{1cm}}{\bf Top-1 Acc.} &\multicolumn{1}{p{1.4cm}}{\bf Throughput (image/s)} \\
\midrule
MaiT-XS    &24.5      &5.3    &81.8   &1853.8 \\
Swin-T     &29     &4.5    &81.3   &1500.8 \\
H-MaiT-T   &29      &4.5    &81.7   &- \\
MaiT-S     &43.3      &9.1   &82.8   &1382.6 \\
Swin-S     &50     &8.7    &83.0   &939.8 \\
H-MaiT-S   &50      &8.9    &83.2   &- \\

%\bottomrule
\end{tabular}
%}
\end{sc}
\end{small}
\end{center}
\vskip -0.1in
\end{table}

We intentionally apply Hard Masking to all heads for the first two stages of the hierarchical structure since this leads to the most meaningful computation saving. Due to smaller patch sizes (4$\times$4 and 8$\times$8), calculating attention maps consumes around 84.5\% and 40.5\% of total FLOPs in the entire transformer block in the first two stages respectively. Applying the attention mask effectively reduces the attention map computation by 99.7\% and 98.9\% for Stage 1 and Stage 2 respectively. As a result, total FLOPs in the first two stages reduce by around 84.2\% and 40\%. 

Meanwhile, with this aggressive masking in the first two stages, H-MaiT still achieves superior accuracy than Swin. Thus H-MaiT offers a novel approach to scale the hierarchical transformer architecture with an efficient reduction in attention map computation. Additionally, this also provides insights on the importance of locality in the early stage of the hierarchical transformer structures.  

\subsection{Embedded Applications}
Although vision transformers have achieved competitive results and even outperform CNN based models in the large parameter domain, performances of vision transformers in the small parameter domain still lag behind CNN based models. This hinders its wide deployment in embedded devices. Though there are various approaches to incorporate locality into transformers, only a few consider the embedded or mobile applications \cite{mobilevit, litevit}. Besides, different from most existing approaches, MaiT is pure transformer based model. This presents a unique hardware acceleration opportunity for NLP and CV unified transformer models. 

With Soft Masking to introduce some spatial locality, distilled MaiT-XT$\dagger$ outperforms EfficientNet-B0 \cite{efficientnet} by 0.4\% with similar parameters as shown in Tabel~\ref{small models}. Moreover, distilled MaiT-XT$\dagger$ achieves 0.9\% higher accuracy than distilled DeiT$\dagger$ with a similar number of parameters and FLOPs, but 600 fewer training epochs. 
%On the other hand, we observe a significant performance improvement with MaiT especially in the smaller model size regime as shown in . Though MaiT consumes more FLOPs, but its throughput is comparable to EfficientNet-B0 \cite{efficientnet}. This shows that the efficiency of pure-transformer based models can match CNN based model in the small parameter regime as well. 

\begin{table}[t]
\caption{Model performance comparison of EfficientNet-B0 \cite{efficientnet}, DeiT-T$\dagger$ \cite{deit}, and MaiT-XT$\dagger$. DeiT-T$\dagger$ is trained with distillation for 1000 epochs. MaiT-XT$\dagger$ is obtained through distillation following the same training process as DeiT$\dagger$ for 400 epochs.}
\label{small models}
\vskip 0.1in
\begin{center}
\begin{small}
\begin{sc}
%\resizebox{\columnwidth}{!}{
\begin{tabular}{llll}
%\multicolumn{1}{l}{\bf}  &\multicolumn{1}{l}{\bf}  \multicolumn{3}{c}{\bf MaiT} \\
\multicolumn{1}{c}{\bf } &\multicolumn{1}{p{1.2cm}}{\bf Param. (M)} &\multicolumn{1}{p{1.2cm}}{\bf FLOPs (G)} &\multicolumn{1}{p{1cm}}{\bf Top-1 Acc.} %&\multicolumn{1}{c}{\bf Throughput} 
\\
\midrule
EfficientNet-B0     &5.3     &0.39    &77.1   \\%&2694.3 \\
%MobileNetV2         &6.9     &0.59    &74.7   & \\
DeiT-T$\dagger$            &6      &1.3    &76.6   \\%&? 
MaiT-XT$\dagger$            &6.4      &1.5    &77.5   \\%&? 
%MaiT-T$\dagger$              &5.7M      &1.25G   &76.6   &2536.5
%\bottomrule
\end{tabular}
%}
\end{sc}
\end{small}
\end{center}
\vskip -0.1in
\end{table}

\subsection{Impact of the Attention Mask}
Training deeper transformer models is difficult because naively stacking transformer layers fails to deliver the expected performance gain \cite{cait}. 
One reason is attention collapse, i.e. the attention maps are more alike among deeper layers \citep{deepvit}. We find mixed masking is able to break the structural repetition in deep transformers and promotes diversity in later layers. 

\begin{figure}[ht]
\vskip 0.1in
\begin{center}
\centerline{\includegraphics[width=\columnwidth]{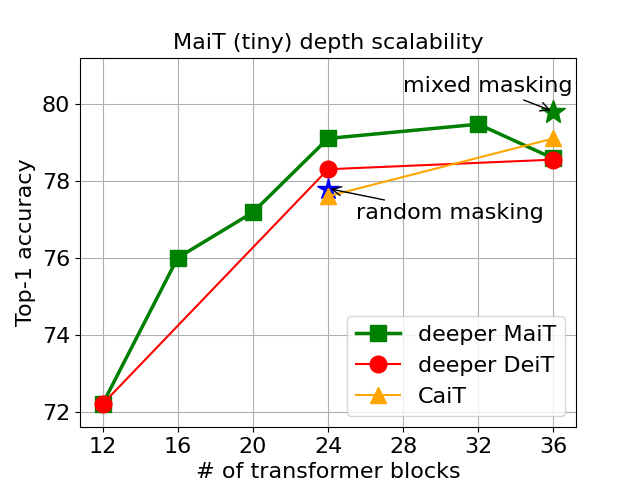}}
\caption{Top-1 accuracy on ImageNet for DeiT-T (red), MaiT-T (green), CaiT-T (yellow), and randomly masked DeiT (blue) with various numbers of transformer blocks. Green square represents MaiT with the same one masked head for all layers. Green star denotes MaiT with mixed masking: three masked heads in the first 12 layers and one masked head for the rest 24 layers.}
\label{scalability}
\end{center}
\vskip -0.1in
\end{figure}

As shown in Figure~\ref{scalability}, top-1 accuracy of DeiT starts to saturate after reaching a certain depth. For example, increasing the number of layers from 24 to 36, top-1 accuracy only increases by 0.25\% to 78.6\% for DeiT-T. In contrast, MaiT with mixed masking boosts the accuracy to 79.8\%, extending the performance gain with deeper transformers. 
Interestingly, we noticed that for 24-layer models, CaiT is less accurate than naively stacked DeiT-24, suggesting that the hyper-parameters for 24-layer CaiT-tiny are probably not optimal. 

To rule out the pure pruning effects on attention maps from masking, we also apply a random mask of the same drop-out rate to one head of all 24 layers. The randomly masked model leads to 0.5\% accuracy drop compared to 24-layer DeiT. This provides further proof that the performance gain with attention masks is indeed from better spatial locality aggregation.  

To better understand the impact of attention masks in deep transformers, we analyze the cross-layer similarity heat map among 36 layers of DeiT, MaiT (one masked head), and MaiT (mixed masking) in the appendix (Figure~\ref{deit similarity}, Figure~\ref{mait similarity}, and Figure~\ref{similarity_mixed}). Mixed attention masking effectively lowers the cross-layer similarity and promotes diverse attention maps at the later layers. As a result, MaiT improves the scalability of vision transformers in the depth direction.

\begin{table*}[ht]
\vskip 0.1in
\caption{Top-1 accuracy for deep MaiT and some SOTA vision transformer models. *Added LayerScale on MaiT. Hyperparameters for MaiT are listed in Appendix Table~\ref{masking}.}
\label{MaiT-deep}
\begin{center}
\begin{tabular}{lllll}
\multicolumn{1}{l}{\bf Model } &\multicolumn{1}{c}{\bf Layers} &\multicolumn{1}{c}{\bf Params(M)} &\multicolumn{1}{l}{\bf GFLOPs} &\multicolumn{1}{l}{\bf Top-1 Acc.} %&\multicolumn{1}{l}{\bf Throughput (im/s)}
\\ \hline
%DeiT-T     &12       &5.7    &1.3    &72.2 &2536\\  
%\textbf{MaiT-T}     &12       &5.7   &1.3    &\textbf{74.7} &? \\ 
LocalViT-T \citep{localvit}      &12    &5.9      &1.3  &74.8 \\
\textbf{MaiT-XT*}     &24       &6.3   &1.5    &\textbf{75.5} \\  
%\textbf{MaiT-XT$^\dagger$}     &24       &6.4   &1.5    &\textbf{76.9} \\  

 \hline
PVT-T \citep{pvt}       &8    &13.2    &1.9    &75.1\\
CaiT-T \citep{cait}      &26    &12    &2.5    &77.6\\
Masked CaiT       &26    &12    &2.5     &78.1\\
%\textbf{MaiT-T}       &24      &11    &2.5     &\textbf{79.1}\\
\textbf{MaiT-T}       &24      &11    &2.5     &\textbf{79.3}\\
\hline
%DeiT-S      &12      &22    &4.6     &79.8  &940\\
%\textbf{MaiT-S}     &12       &22   &4.6    &\textbf{80.1} \\ 
%tiny    &CaiT*      &24+2    &12    &2.5    &?\\
%\textbf{MaiT-XS}     &24      &24.5    &5.3  &\textbf{81.4}  \\
PVT-S \citep{pvt}      &15    &24.5    &3.8    &79.8\\
LocalViT-S \citep{localvit}     &12    &22.4      &4.6  &80.8 \\
Swin-T \citep{swin}          &12    &29        &4.5  &81.3 \\ %&755.2
T2T-ViT-14 \citep{t2t}     &16    &21.5      &5.2  &81.5 \\
Twins-SVT-S \citep{twins}      &18      &24       &2.8    &81.7\\
\textbf{H-MaiT-T}     &12      &29    &4.5  &\textbf{81.7}  \\
CaiT-XS \citep{cait}     &26    &26.6    &5.4  &81.8 \\
%DeepViT-S     &16    &27    &6.2  &82.3 \\
\textbf{MaiT-XS*}     &24      &24.5    &5.3  &\textbf{81.8}  \\
%\textbf{Masked CaiT-XS}     &24+2    &26.6    &5.4  &\textbf{?} \\
\hline
%Extra Small    &CaiT*      &24+2    &26.6    &5.4  &? \\
%DeiT-S$^+$      &24      &43.3      &9.1    &81.0 \\
%\textbf{MaiT-S}      &24      &43.3    &9.1     &\textbf{82.1} \\
PVT-M \citep{pvt}      &27    &44.2    &6.7    &81.2\\
T2T-ViT-24 \citep{t2t}     &26    &64.1      &14.1  &82.3 \\
CaiT-S \citep{cait}     &26    &46.9      &9.4  &82.7 \\
Masked CaiT-S      &26      &46.9    &9.4    &82.9 \\
\textbf{MaiT-S*}      &24      &43.3    &9.1    &\textbf{82.9} \\
Swin-S \citep{swin}     &24    &50      &8.7  &83.0 \\ %&436.9
Twins-SVT-B \citep{twins}      &24      &56       &8.3    &83.2 \\
\textbf{H-MaiT-S}     &24      &50    &8.9  &\textbf{83.2}  \\
%DeepViT-L     &32    &58    &12.8  &83.1 \\

%Small    &CaiT*      &24+2    &46.9      &9.4  &? \\
\end{tabular}
\end{center}
\vskip -0.1in
\end{table*}

\subsection{Comparison with other SOTA vision transformers}
Attention masks can be easily combined with other optimization techniques. For deep transformers, we adopt the LayerScale \citep{cait} to MaiT. As shown in Table~\ref{MaiT-deep}, MaiT obtains competitive performances across different model sizes compared to some of the SOTA transformer models. Specifically, MaiT-T and MaiT-S outperform CaiT-T and CaiT-S by 1.7\% and 0.2\% respectively with fewer parameters/FLOPs and simpler structure, which translates to higher throughput. Moreover, simply adding the attention mask to CaiT leads to 0.5\% and 0.2\% higher accuracy for tiny and small models respectively. In addition, for the hierarchical structure, H-MaiT-T and H-MaiT-S achieves 0.4\% and 0.2\% higher accuracy than Swin-T and Swin-S respectively. H-MaiT-T and H-MaiT-S obtain the same accuracy as Twins-SVT-S and Twins-SVT-B with a similar number of parameters and FLOPs. 

\section{Conclusion}
In this work, we provide an effective model-agnostic approach with attention masks to improve the scalability of vision transformers. Masked attention heads efficiently reduce the complexity to compute attention map from quadratic to linear with respect to the number of input tokens. %without sacrificing accuracy or throughput.
Additionally, local and global dependencies are captured simultaneous with masked and unmasked attention heads respectively. This improves the efficiency and performances of pure-transformer based models, especially at the small model size regime for embedded applications. Moreover, it is promising to further boost performances by combining this attention mask  technique with other optimization approaches. 
%In this work, we incorporate spatial locality into vision transformers by inserting masks to attention heads.Masked heads are able to focus on local information and liberate other unmasked heads to extract global information more effectively. We also introduce attention locality score to guide the mask search and rapidly evaluate various masking schemes. Attention masking is a simple and effective technique, especially for small and deep transformer models. We observe that attention masks also serve as a regularizer to guide the training of attention maps for deeper transformers. 

%\section*{Acknowledgements}


\bibliography{references}
\bibliographystyle{icml2022}

%%%%%%%%%%%%%%%%%%%%%%%%%%%%%%%%%%%%%%%%%%%%%%%%%%%%%%%%%%%%%%%%%%%%%%%%%%%%%%%
%%%%%%%%%%%%%%%%%%%%%%%%%%%%%%%%%%%%%%%%%%%%%%%%%%%%%%%%%%%%%%%%%%%%%%%%%%%%%%%
% APPENDIX
%%%%%%%%%%%%%%%%%%%%%%%%%%%%%%%%%%%%%%%%%%%%%%%%%%%%%%%%%%%%%%%%%%%%%%%%%%%%%%%
%%%%%%%%%%%%%%%%%%%%%%%%%%%%%%%%%%%%%%%%%%%%%%%%%%%%%%%%%%%%%%%%%%%%%%%%%%%%%%%
\newpage
\appendix
\onecolumn
\section{Appendix}

%You can have as much text here as you want. The main body must be at most $8$ pages long.
%For the final version, one more page can be added.
%If you want, you can use an appendix like this one, even using the one-column format.
%%%%%%%%%%%%%%%%%%%%%%%%%%%%%%%%%%%%%%%%%%%%%%%%%%%%%%%%%%%%%%%%%%%%%%%%%%%%%%%
%%%%%%%%%%%%%%%%%%%%%%%%%%%%%%%%%%%%%%%%%%%%%%%%%%%%%%%%%%%%%%%%%%%%%%%%%%%%%%%

% This document was modified from the file originally made available by
% Pat Langley and Andrea Danyluk for ICML-2K. This version was created
% by Iain Murray in 2018, and modified by Alexandre Bouchard in
% 2019 and 2021 and by Csaba Szepesvari, Gang Niu and Sivan Sabato in 2022. 
% Previous contributors include Dan Roy, Lise Getoor and Tobias
% Scheffer, which was slightly modified from the 2010 version by
% Thorsten Joachims & Johannes Fuernkranz, slightly modified from the
% 2009 version by Kiri Wagstaff and Sam Roweis's 2008 version, which is
% slightly modified from Prasad Tadepalli's 2007 version which is a
% lightly changed version of the previous year's version by Andrew
% Moore, which was in turn edited from those of Kristian Kersting and
% Codrina Lauth. Alex Smola contributed to the algorithmic style files.

\subsection{Hyper parameters for MaiT}

\begin{table}[h]
\caption{Hyper parameters for 24-layer MaiT in Table~\ref{MaiT-deep}. Drop path rate refers to linear stochastic depth drop rate, the same as in Trouvron et al. \yrcite{deit}. *denotes a constant drop path rate for all layers as in Trouvron et al. \yrcite{cait}.}
\label{masking}
\begin{center}
\begin{tabular}{llll}
\multicolumn{1}{l}{\bf Model}  &\multicolumn{1}{c}{\bf Drop path rate}  &\multicolumn{1}{c}{\bf Batch size} &\multicolumn{1}{c}{\bf Learning rate}  
\\ \hline \\
%MaiT-XT     &3/0-7 + 1/8-21      &0.0   &1024   &0.001\times batchsize/512\\
%MaiT-T      &1/0-23              &0.05   &1024   &0.001 \times batchsize/512\\
%MaiT-XS     &1/0-23              &0.1   &1024    &0.0005 \times batchsize/512\\
%MaiT-S      &5/0-7 + 1/8-19      &0.1*   &1024   &0.0005 \times batchsize/512\\
MaiT-XT      &0.0   &1024   &0.001$\times$batchsize/512\\
MaiT-T       &0.0   &1024   &0.001$\times$batchsize/512\\
MaiT-XS      &0.1   &1024    &0.0005$\times$batchsize/512\\
MaiT-S       &0.1*   &1024   &0.0005$\times$batchsize/512\\
\end{tabular}%
\end{center}
\end{table}

\subsection{Cross-layer similarity}

To evaluate the impact of attention masks on the diversity of the attention maps across all layers, we apply a cross-layer similarity metric similarly defined in DeepViT \citep{deepvit}:

\begin{equation} \label{similarity}
      M^{i, j}_{h,t} =  \frac {A^{i}_{h, t, :} A_{h, t, :}^{j \top}}
    {\|A^{i}_{h, t, :} \| \|A^{i}_{h, t, :} \|}  
\end{equation}
where $M^{i, j}_{h,t}$ is the attention map cosine similarity between layer i and layer j for attention head h and token t. $A_{h, t, :}$ measures the weight contribution from each token in input (Value) for output token T. Therefore, it reflects the similarity of attention maps on information aggregation across all T input tokens. For example, if $M^{i, j}_{h,t}$ reaches 1, output token t at head h attends to all N+1 tokens in Value with the same probability at both Layer i and Layer j. 

\begin{figure}[ht]
\begin{center}
    \includegraphics[scale=0.46]{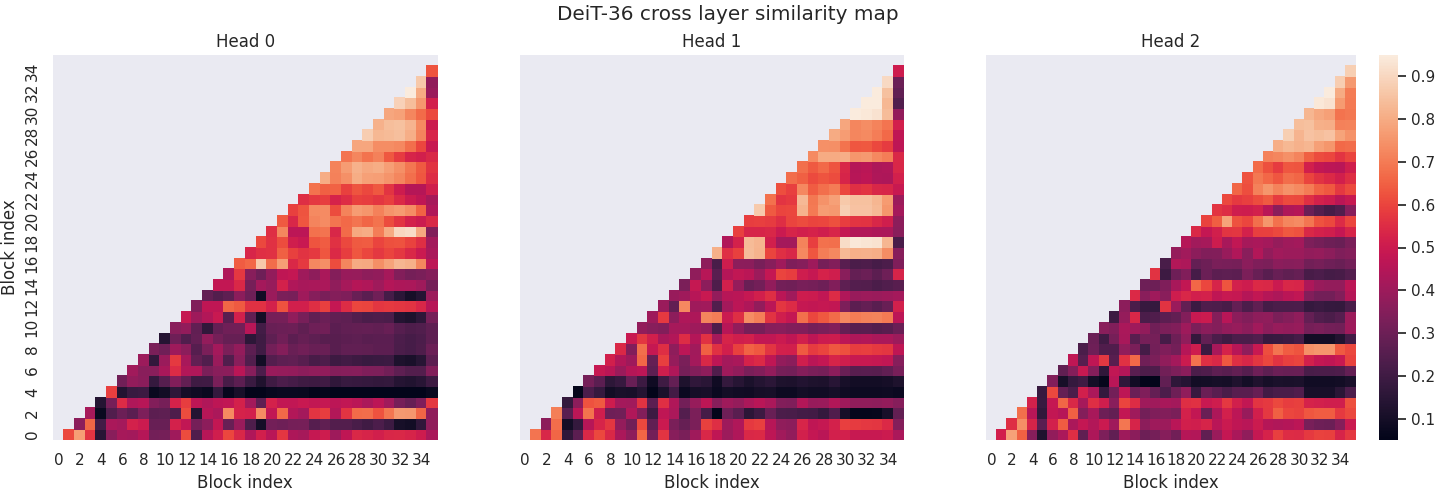}
\end{center}
\caption{Attention map similarity across 36 layers for DeiT-T}
\label{deit similarity}
\end{figure}

We analyze the cross-layer similarity of 36-layer DeiT-T and MaiT-T to explore the impact of attention masks. DeiT shows the highest cross-layer similarity at the last 8 stages, with an average of 0.75 (Figure~\ref{deit similarity}). This homogeneity in attention maps at the late stage limits the model learning capability. 

\begin{figure}[ht]
\begin{center}
    \includegraphics[scale=0.46]{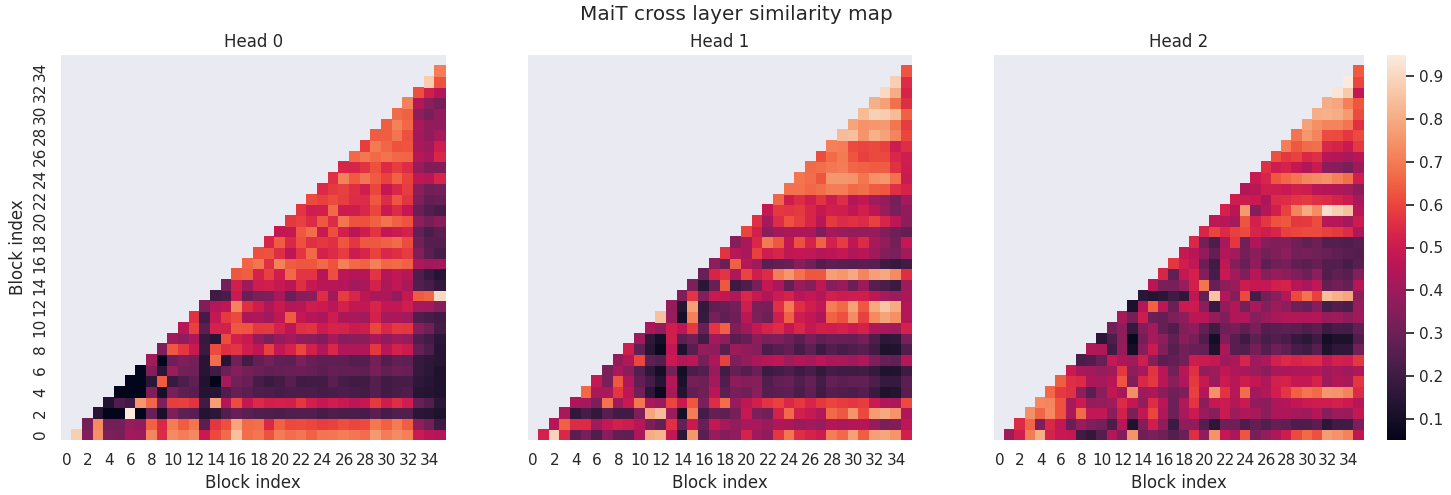}
\end{center}
\caption{Attention map similarity across 36 layers for MaiT-T with one masked attention head.}
\label{mait similarity}
\end{figure}

Adding attention mask to the same one head for all 36 layers does not alleviate this problem since it's still the same structure repeating 36 times along the depth direction. As a result, the average cross-layer similarity for the last 8 layers with unmasked heads is also 0.75 (Figure~\ref{mait similarity}). 
This suggests one masked head for all transformer blocks is not effective in 36-layer models. Repeating the same structure for 36 times in the depth direction is causing the attention collapse \cite{deepvit}. The diversity among masked and unmasked heads within the same layer doesn't alleviate this issue. 

\begin{figure}[h]
\begin{center}
    \includegraphics[scale=0.46]{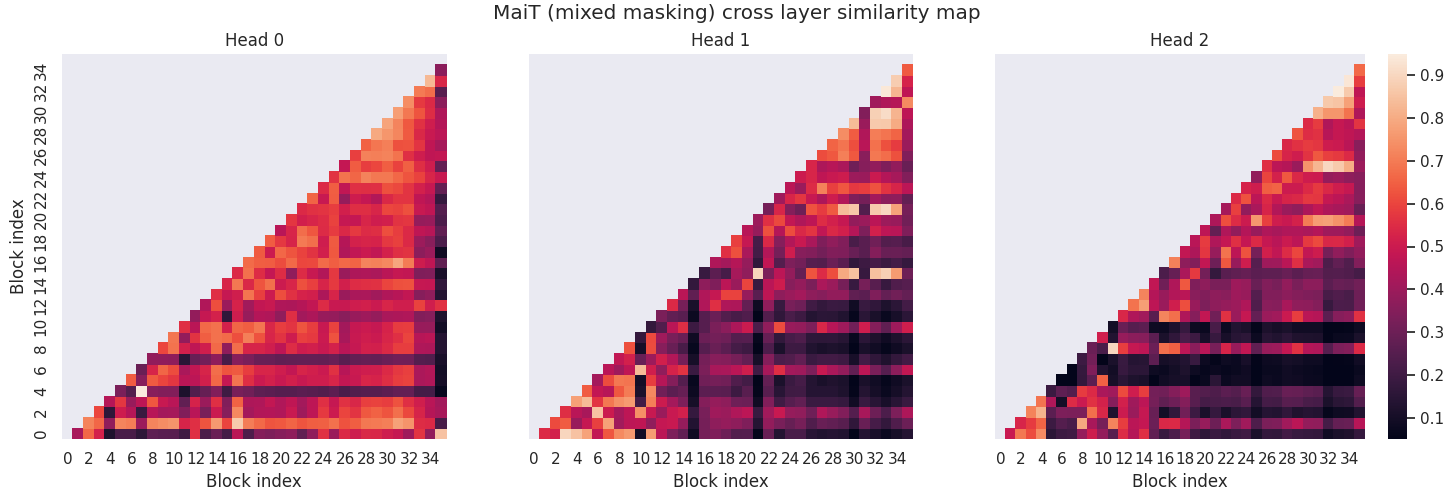}
\end{center}
\caption{Attention map similarity across 36 layers for MaiT-T with mixed masking scheme}
\label{similarity_mixed}
\end{figure}

Alternatively, with mixed masking scheme, all masked heads in the first 8 layers are naturally different from the rest of the partially masked layers, and the same structure only repeat 8 or 24 times. As shown in Figure~\ref{similarity_mixed}, cross-layer similarity between the first 8 layers and the following 24 layer are mostly well below 0.3 for Head 1 and Head 2. Moreover, a mixed masking scheme lower the cross-layer similarity among the last 8 layers by 0.1 to 0.65. This indicates mixed masking scheme effective increases the cross layer diversity especially in the last few layers, which improves model performances.  

%In Figure~\ref{deit similarity}, 36-layer DeiT-tiny shows the highest similarity at the last 8 stages across all three heads, with an average similarity of 0.75, whereas the similarity is quite small among the first 16 layers and between the first 16 layers and the last 20 layers. This is agreement with \cite{deepvit}, indicating an attention collapse at later stage of deep transformers.  

\end{document}